# A proposed method to extract maximum possible power in the shortest time on solar PV arrays under partial shadings using metaheuristic algorithms

Reza Hedayati Majdabadi and Saeed Sharifian Khortoomi

Department of Electrical Engineering, Amirkabir University of Technology, 15914, Tehran, Iran.

{rhedayati, sharifian_s}@aut.ac.ir

**Abstract:** The increasing use of fossil fuels to produce energy is leading to environmental problems. Hence, it has led the human society to move towards the use of renewable energies, including solar energy. In recent years, one of the most popular methods to gain energy is using photovoltaic arrays to produce solar energy. Skyscrapers and different weather conditions cause shadings on these PV arrays, which leads to less power generation. Various methods such as TCT and Sudoku patterns have been proposed to improve power generation for partial shading PV arrays, but these methods have some problems such as not generating maximum power and being designed for a specific dimension of PV arrays. Therefore, we proposed a metaheuristic algorithm-based approach to extract maximum possible power in the shortest possible time. In this paper, five algorithms which have proper results in most of the searching problems are chosen from different groups of metaheuristic algorithms. Also, four different standard shading patterns are used for more realistic analysis. Results show that the proposed method achieves better results in maximum power generation compared to TCT arrangement (18.53%) and Sudoku arrangement (4.93%). Also, the results show that GWO is the fastest metaheuristic algorithm to reach maximum output power in PV arrays under partial shading condition. Thus, the authors believe that by using metaheuristic algorithms, an efficient, reliable, and fast solution is reached to solve partial shading PV arrays problem.

## I. INTRODUCTION

Nowadays, the importance of energy, especially electrical energy is known to everyone. Shortage of energy and air pollution caused by fossil fuels are the main problems that threaten human life. Due to the limitation of fossil fuels, the use of new energy sources is increasingly being considered. New energy sources are growing fast and getting a greater share of the energy market every day. According to the reports of the International Energy Agency, the gap between production and demand for non-renewable energies is growing day by day, and they predicted that it would reach its critical point in 2030. Thus, this is the reason to think about an efficient way to overcome these issues.

In recent years, the development of renewable energies such as solar photovoltaic power generation and wind turbine has been a significant contribution to provide the required energy [1]. According to the International Renewable Energy Agency, the share of solar power in energy consumption by the year 2016 was about 2 percent, and it will get to 13 percent in 2030.

Solar photovoltaic cells are able to convert solar power to electrical power in the range of 50 to 350 watts. Photovoltaic cells technology is based on silicon, so development in semiconductors technology has enabled us to get power from photovoltaic cells as much as the power plant [2]. Another benefit of using this type of photovoltaic cells is that, unlike concentrating solar power cells, they do not need to be irradiated directly from the sun. They can produce electrical energy even when they are shaded. As a result, using these types of cells will be a good alternative solution for the future.

As mentioned in [3], the low power of solar cells alone makes it possible to connect them through different interconnections to achieve more power. The solar cells are connected in series to form PV modules. PV arrays are formed by connecting the PV modules in series and parallel. There are multiple methods to make PV arrays such as series-parallel, Total-cross-Tied, Bridge-Link and Honey Comb that are discussed in detail in [4]-[5]. One of the main challenges of using PV-arrays is fall of the shadow on the cells that results in decreasing of output power. The shading on the cells is not only the reason for the decreased performance of solar cells but also how cells are connected, shading pattern and the amount of shading are effective. As mentioned in [6], TCT configuration has the best performance under different shading patterns. One of the TCT configuration problems is peaks in the Power-Voltage diagrams in different PV patterns. Several methods have been proposed to solve TCT configuration problems. For instance, Rani et al. [7] achieve more power than TCT arrangement under different shading patterns without changing electrical connections, and they also solve the peaks problem in the PV diagram. However, this method is can only be used in 9x9 cells. Therefore, in 2015, Namani et al. [8] proposed a method using a magic square to achieve more power than TCT configuration by changing the location of cells without changing the electrical connections of the cells. Also, Pachauri et al. [9] using this method and Latin square puzzle pattern to reduce power losses and to improve the performance of the TCT configuration. Thus, by changing the location of cells physically, the shadows distribute appropriately through PV arrays. In 2015, Deshkar et al. [10] proposed a method by changing electrical connections using Genetic algorithm to spread the shades on solar cells to achieve more power and softer diagram with less local peaks. The performance of this method shows better results than recent best methods such as Sudoku and different TCT configuration.

Using metaheuristic algorithms for solving different types of continuous and discrete search and optimization problems had remarkable growth over the last years. The main reason that these algorithms are so popular in scientific society is their easy implementation and the short time needed to reach the best answer. Therefore, the authors intend to find an algorithm to reach the maximum possible power in minimum time in different shading patterns on solar arrays. In this paper, five biological and evolutionary algorithms, that each one is in a different metaheuristic algorithm family, will be discussed. These algorithms are a Genetic Algorithm (GA), Multiverse Optimization (MVO), Moth and Flame Optimization (MFO), Imperial Competition Algorithm (ICA) and Gray Wolf Group Optimization (GWO) algorithm.

## II. DISCUSSED SYSTEM DESCRIPTION

In this section, the electronic model of solar PV arrays is introduced. Then, the proposed model, standard shading patterns, and proposed metaheuristic algorithms are described briefly.

### A. *Proposed structure*

Different structures have been proposed for solar cells, that are described in [11] - [14]. The structure considered in this paper is composed of a current source and a diode connected in parallel. The received light is directly proportional to generated current from the current source calling photocurrent. When there is no light on the solar cell, the current source will not generate any current. Therefore, the diode is biased in the reverse region. In other words, the diode in this circuit is the representation of I-V diagram of the solar cell. Electronic diagram of this circuit is shown in figure 1. [15]

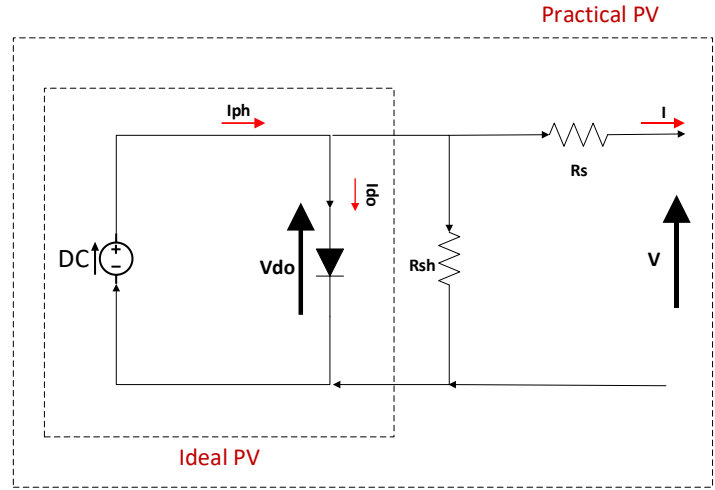

**Figure 1. Equivalent circuit of a solar PV cell**

By considering circuit rules output current (*I*) and voltage (*V*) can be defined by equations 1 to 3.

$$I_{do} = I_0(\exp(\frac{qV_{do}}{n.kT}) - 1) \tag{1}$$

$$I = I_{ph} - I_{do} - \frac{V_{do}}{R_{sh}} \tag{2}$$

$$V = V_{do} - R_s I \tag{3}$$

In equation 1, $I_{do}$ is the mean current passing the diode, $I_0$ is the reverse saturated current, $n$ is a parameter obtained from diode characteristic, $q$ is the electron charge, $k$ is the Boltzmann constant and $T$ is the temperature of the solar cell. In equation 2, $I_{ph}$ is the photocurrent. As you can see in equations 1-3 and figure 1, two parallel and series resistors, $R_s$ and $R_{sh}$ are in the circuit which represent nonconductive and current leakage of weak insulation respectively. In ideal state, these resistances are $R_s = 0, R_{sh} = \infty$. It can be concluded from equations 1 to 3 that the output current of circuit is:

$$I = I_{ph} - I_0(\exp(\frac{q}{n.kT}(V + R_s I)) - 1) - (\frac{V + R_s I}{R_{sh}}) \tag{4}$$

By ignoring the series resistance in the ideal state, the final output current of the circuit can be defined by equation 5.

$$I = I_{ph} - I_0(\exp(\frac{q}{n.kT}V) - 1). \tag{5}$$

$I_{ph}$ can be defined by equation 6.

$$I_{ph} = I_{sc} \left(\frac{G}{G_0}\right)(1+\alpha_1(T-T_0))\frac{R_s+R_{sh}}{R_{sh}} \qquad (6)$$

As you can see in equation 6, $I_{sc}$ is used for calculating photocurrent. Short circuit current can be defined by equation 7:

$$I_{sc} = I = \frac{I_{ph}}{(1+\frac{R_s}{R_{sh}})} \qquad (7)$$

Open circuit voltage can be defined by equation 8:

$$V_{oc} \approx \left(\frac{n.k\,T}{q}\right)\ln\left(\frac{I_{ph}}{I_0}\right) \qquad (8)$$

At last, the solar cell power is calculated by using the equation P=VI.

$$P = (I_{ph} - I_{do} - \frac{V_{do}}{R_{sh}})V \qquad (9)$$

The parameters mentioned above are defined in table 1.

Table 1. Parameter Definition

| Parameter | Definition |
|---|---|
| $G$ | Cell Irradiance |
| $G_0$ | Standard Irradiance |
| $I$ | Cell Current |
| $I_0$ | Reverse Saturation Current |
| $I_{do}$ | Average Diode Current |
| $I_{ph}$ | Photoelectric Current |
| $I_{sc}$ | Short Circuit Current |
| $k$ | Boltzmann's Constant |
| $n$ | Number of Series Connected Cells |
| $P$ | Cell Power |
| $q$ | Electric Charge |
| $R_{sh}$ | Shunt Resistance |
| $R_s$ | Series Resistance |
| $T$ | Cell Temperature |
| $T_0$ | Standard Temperature |
| $V$ | Cell Voltage |
| $V_{do}$ | Diode Voltage |
| $V_{oc}$ | Open Circuit Voltage |
| $\alpha_1$ | Module's Temperature Coefficient |

## B. *Proposed Model*

As mentioned before, each PV array is constructed by putting together n×m PV modules leading to a n×m matrix. These modules connect in a different way noted in section 1, one of which is series-parallel. In this arrangement, all modules in a column connected in series and all series connected in parallel. TCT pattern, inspired by the series-parallel pattern, in addition to series connections of all modules in each column, all modules in each row connected in parallel. For more clarity this pattern is shown in figure 2.

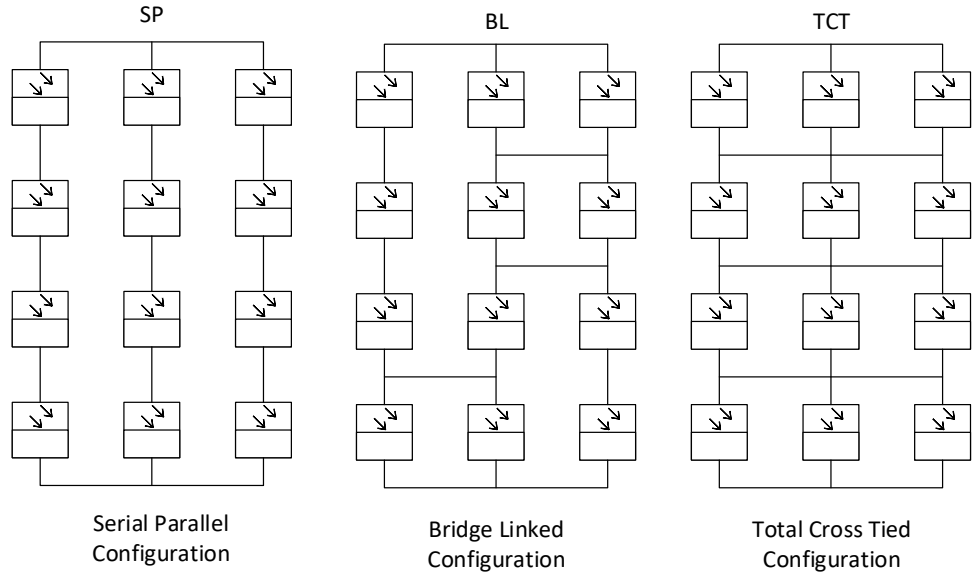

**Figure 2. Different PV arrays topologies**

The fixed physical arrangement of TCT pattern causes the modules cannot rearrange in shading environments. Therefore, it prevents them from gaining maximum output power. Moreover, if there is heavy shading on a row, that leads to a significant decrease in row current, the row should be bypassed from the circuit to prevent damage. This results in unwanted peaks in I-V diagram. In order to overcome these problems, it is sufficient to change modules location physically without changing their electrical connections. Thus, the shadow uniformly is distributed all over the array by replacing modules in its columns. This idea was used in [16] by using Sudoku algorithm for the first time.

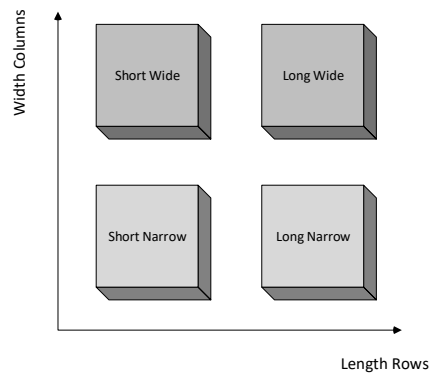

**Figure 3. Different types of shading patterns**

The results yielding from Sudoku algorithm show that although total power has a notifiable improvement concerning the TCT pattern, it is still far from its maximum value. Furthermore, physical rearrangements of modules are costly. Therefore, for overcoming these issues, metaheuristic algorithms are used.

In General, different shading pattern including short-wide, long-wide, short-narrow and long-narrow, which were introduced in [17], are used. These patterns are shown in figure 3.

In the following, the metaheuristic algorithms which is used in this paper to change electrical connections of solar cells without physical rearrangement are introduced. The goal of using these algorithms is to solve previous methods problems. Because of unique features of metaheuristic algorithms, it is predictable to have fast convergence with the maximum possible power even when there is partial shading on PV arrays.

### C. *Metaheuristic Algorithms*

Metaheuristic algorithms are inspired by natural phenomena, which search intelligently for the best answer in a wide diversity of complicated problems. In other words, intelligence means instead of searching all possible solutions that may contain many non-optimal results, searching the areas that are more probable to have an acceptable answer.

The goal of this paper is to find a way to generate maximum possible power in the shortest possible time for partial shading PV arrays. Replacing each module in the PV array, any of which is a unique solution causes a large search space. Hence, it is better to use metaheuristic algorithms to obtain maximum possible power in the shortest possible time.

Metaheuristic algorithms, as the most intelligent way of searching methods, inspires from a natural heuristic for gauging problems with rough landscapes. These algorithms are divided into two categories: biological and non-biological. Biological methods are also divided into evolutionary and metaheuristic algorithms based on crowd intelligence. Non-biological methods are also categorized into two groups: physic-based algorithms and other non-biological metaheuristic methods. In this paper, the authors intend to select the most suitable algorithms from each group of metaheuristic algorithms in order to find the best solution to the problem.

Evolutionary algorithms are a kind of metaheuristic algorithms based on survival of the fittest theory by Darwin. The Genetic algorithm (GA) [18] that is the most common method in this group is chosen. This algorithm was used in many solar power problems like optimizing for sun tracker trajectory [19]. Metaheuristic algorithms based on crowd intelligence are indirectly relevant to Darwin's theory and inspired by social creatures' behaviors. The Imperial Competition Algorithm (ICA) [20] and Gray Wolf Group Optimization (GWO) are chosen [21] from this group because of their great performance. Non-biological methods based on physics are commonly inspired by a physical law or phenomenon. Multiverse Optimization (MVO) [22] algorithm is chosen from this group. At last, from non-biological non-physics based metaheuristic methods, Moth and Flame Optimization (MFO) [23] algorithm is chosen.

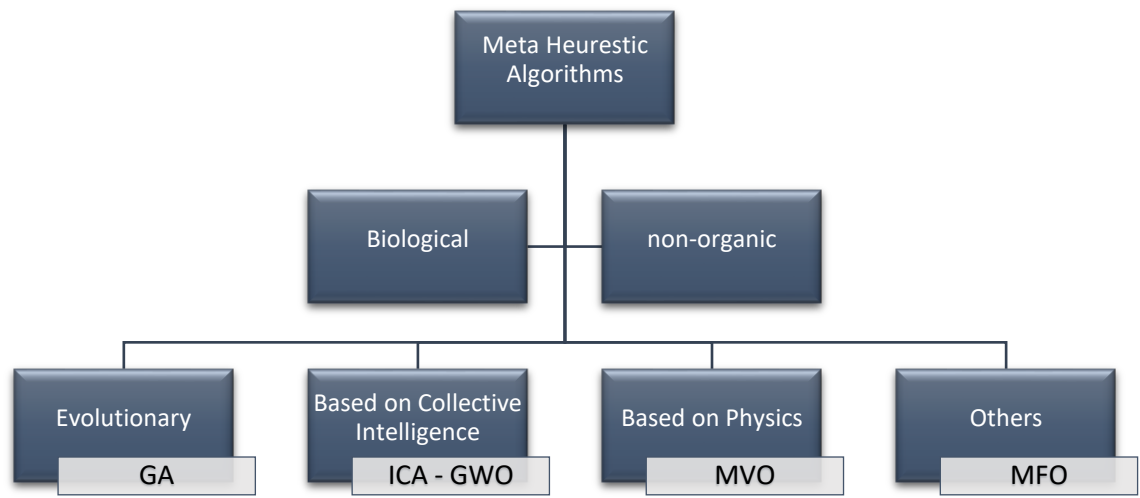

**Figure 4. Proposed meta-heuristic algorithms**

*D. Genetic Algorithm*

According to Darwin's theory, generations who have superior features and characteristics are more probable to survive and proliferate. Their features and characteristics will be inherited to the next generations. According to the second part of Darwin's theory, when a new generation is proliferated, some random events may affect their characteristics. If the changes cause improvement, the probability of the child's survival will be increased.

The procedure of the Genetic Algorithm is as follows. First, some solutions are created by a random or specific pattern, which is called the primitive population, and each solution is called a chromosome. After that, the best chromosomes are chosen. Then, chromosomes are combined with one another and get mutated by using Genetic Algorithm operators. The best chromosomes are chosen to generate the next level population by checking the problem metrics and fitness function of the given population. Finally, the best chromosome fitting the problem metrics is chosen as the best solution.

*E. Imperial Competition Algorithm*

This algorithm, based on evolutionary computation, seeks the optimal solution for different optimization problems. The ICA algorithm is proposed to optimize the problem by using the mathematical model of a political-social evolution process. Optimization solutions in this algorithm are different countries and solutions become better in a repetitive process until the final optimal solution is found.

Like Genetic Algorithm, ICA starts with a random initial population that each solution is symbolized as a country. Some of the best solutions are chosen to be imperialists (equivalent to elites in the genetic algorithm). Other members of the population are called colonies. Colonies are attracted to imperialists based on their powers. The total power of each empire consists of both imperialist country (as its core) and its colonies. This power is mathematically modeled by summation of the power of imperialist plus a percentage mean of colonies powers.

By formation of first empires, imperials competition starts among them. Each empire which not be able to succeed in this competition (empower or defend itself), will be eliminated from the imperial competition. Therefore, the survival of an empire depends on its power tries to attract other empire's colonies and dominate them. Consequently, during the colonial struggles, the more powerful empires will gradually be increased, and the weaker ones eliminated.

### F. Gray Wolf group Algorithm

The GWO algorithm was proposed by Mirjalili et al. [21] in 2014. This algorithm based on wolves hunting behavior and their sovereignty hierarchy such as hunting, searching for bait, surrounding it and attacking the bait. The goal of this algorithm is to find the best bait (best solution) and moving towards it as wolves move toward their bait. The main procedure of this algorithm is as follows. Like evolutionary algorithms, the first step is creating an initial population (wolves). In each iteration of the algorithm, the best three wolfs that have highest fitness function in the population must be found. The best one is called alpha, the second best is called beta, and the third one is called delta. In other words, these three wolves represent the best possible solutions in the current population. They move towards bait in their neighboring radius, and the remaining wolves move in a direction inspired by them. This procedure continues until they reach the bait.

### G. Multiverse Optimization

Multiverse Optimization Algorithm was proposed by Mirjalili et al. [22] in 2015. This algorithm for parameterizing the problem uses physics rules of multiverse theory. MVO based on the hypothesis that every universe is expanding and the more vast universes are more stable.

At first, some initial multiverses create, each of which is a solution to the problem. The target of this algorithm is to find the most expanded multiverse which has the most value of fitness function. Each multiverse has different features. In each iteration of the algorithm, features from a white hole in one of the best multiverses transfer to a black hole in one of the worst multiverses. Then, features from the best universe are transferred to other multiverses through wormholes by a predefined probability. This loop continues until the most expanded multiverse has no improvement (solution is found).

### H. Moth and Flame Algorithm

This algorithm was proposed by Mirjalili et al. [23] in 2015, which is inspired by the Moth flying path around flames. This path is modeled by mathematical equations used in this algorithm to find new solutions.

This algorithm starts with some flames as an initial population. Around each flame, there is a moth flying. Moths which are flying in the further distance have weaker flames light, and vice versa. If a moth finds a point in the search space that has greater fitness function value than its flame, the point considers as a new flame, and a new moth fly around it. Moth's spiral shape flying path causes to search for more points to find the best locations. In fact, this algorithm is inspired by moth's flying path around the flame as a search path. The main equation governing Moths flying path shown in equation 10.

$$S(M_i, F_j) = D_i e^{bt} \cdot \cos(2\pi t) + F_j \tag{10}$$

Where $D_i$ is the distance between the $i_{th}$ Moth and the $j_{th}$ flame, $F_j$ is the location of $j_{th}$ flame, $b$ is a constant for defining the logarithmic shape of spiral and different values of $t$ gives us different solutions (locations) of the problem.

## III. IMPLEMENTATION

In this section, the implementation of proposed algorithms on the described problem will be discussed. The proposed method will be tested on a PV array under 4 different shading patterns including 81 PV modules in a TCT arrangement as a 9×9 matrix. Every metaheuristic algorithm has two main parameters, the initial population and the fitness function which is defined by problem specifications. Different arrangements of the PV array under partial shading conditions are created by rearranging PV modules in their column. These different arrangements generate population for metaheuristic algorithms.

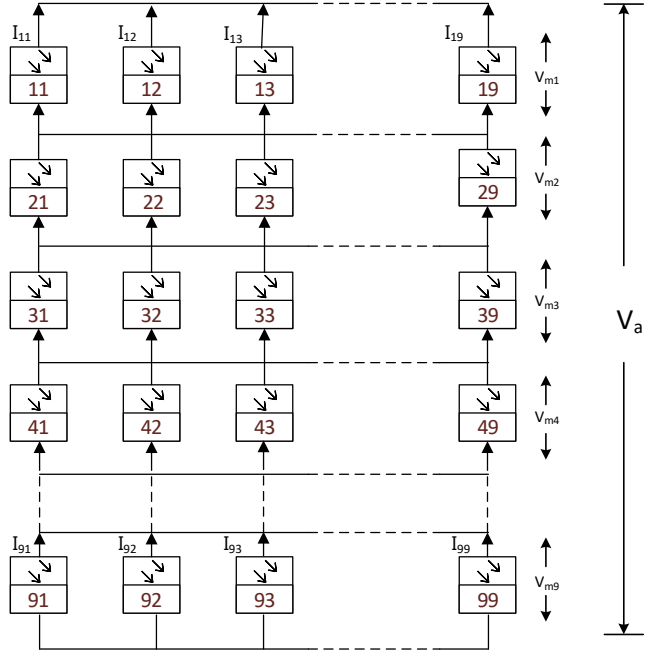

**Figure 5. PV array connected in TCT configuration.**

As mentioned before, the TCT reconfiguration initially is used for this problem. The PV array has $r=9$ rows and $c=9$ columns. $I_{ij}$ and $V_{ij}$ are used for representing current and voltage where '$i$' denotes to the row and '$j$' denotes to the column. The current generated by the module at an irradiance G is given by $I=k_{ij} \times I_m$. Since the panels in a row are connected in parallel, the maximum possible current output of each row is calculated by equation 11.

$$I_i = \sum_{j=1}^{c} k_{i,j} \times I_m \qquad (11)$$

Where $I_m$ is the current produced by each module under standard condition when standard irradiance is $G_0 = 1000 W/m^2$ and temperature is 25 Celsius. $K_{ij}$ is irradiance strength of PV module $i_{th}$ row and $j_{th}$ column derived by $k_{i,j} = \dfrac{G_{i,j}}{G_0}$. It is worthy to note that $I_m$ value and the other PV module specifications at standard test conditions are mentioned in Table 2. Consequently, the current of every row is affected by the received irradiance of that rows' modules. As you can see in figure 5, the current value at each node in the array can be calculated using Kirchhoff's current law.

$$I_{wb} = \sum_{j=1}^{c}(I_{ij} - I_{(i+1)j}) = 0, i = 1,2,3,...,c \tag{12}$$

Also, the array voltage is calculated using KVL. The array voltage is equal to the summation of each rows' voltage.

$$V_{wb} = \sum_{i=1}^{r} V_i \tag{13}$$

Where $V_{wb}$ is the voltage of the PV array and $V_i$ is the voltage of the panels at the $i$th row.

**Table 2. PV Specification at 1000 W/m² and 25°C**

**Table 2**

Specifications for PV at $1000 \ w/m^2$ and 25°C

| | |
|---|---|
| PV Power | 80 W |
| Open Circuit Voltage | 21.24 V |
| Short Circuit Current | 4.74 A |
| Nominal Voltage | 17.64 V |
| Nominal Current | 4.54 A |

The second parameter which is used by metaheuristic algorithm for solving the problem is fitness function. The fitness function is shown in equation 14.

$$Minimize(Fitness(i)) = \frac{MF}{Sum(P) + (\frac{W_t}{C_t}) + (P_{wb} \times W_{wb})} \tag{14}$$

*MF* is a minimization factor in equation 14 for minimizing the fitness function. In dominator of the fitness function, there are three mathematical phrases which are described in following. *Sum(P)* is equal to the summation of voltage and current multiplications of all rows.

$$Sum(P) = \sum_{i=1}^{r} I_i \times V_i \tag{15}$$

The next phrase is the fraction of currents that $W_t$ is the predefined weight for $C_t$ and $C_t$ is derived by equation 15.

$$C_t = \sum_{i=1}^{r} |I_m - I_i| \tag{16}$$

Where $I_m$ is the maximum possible value of current when bypassing is considered. At last the final element is the value of PV array power without bypassing any rows. Predefined $W_{wb}$ weight is used for this parameter. These weights make it possible to control exploration and exploitation in this problem.

In the following, the general process of proposed method is shown in figure 6-7.

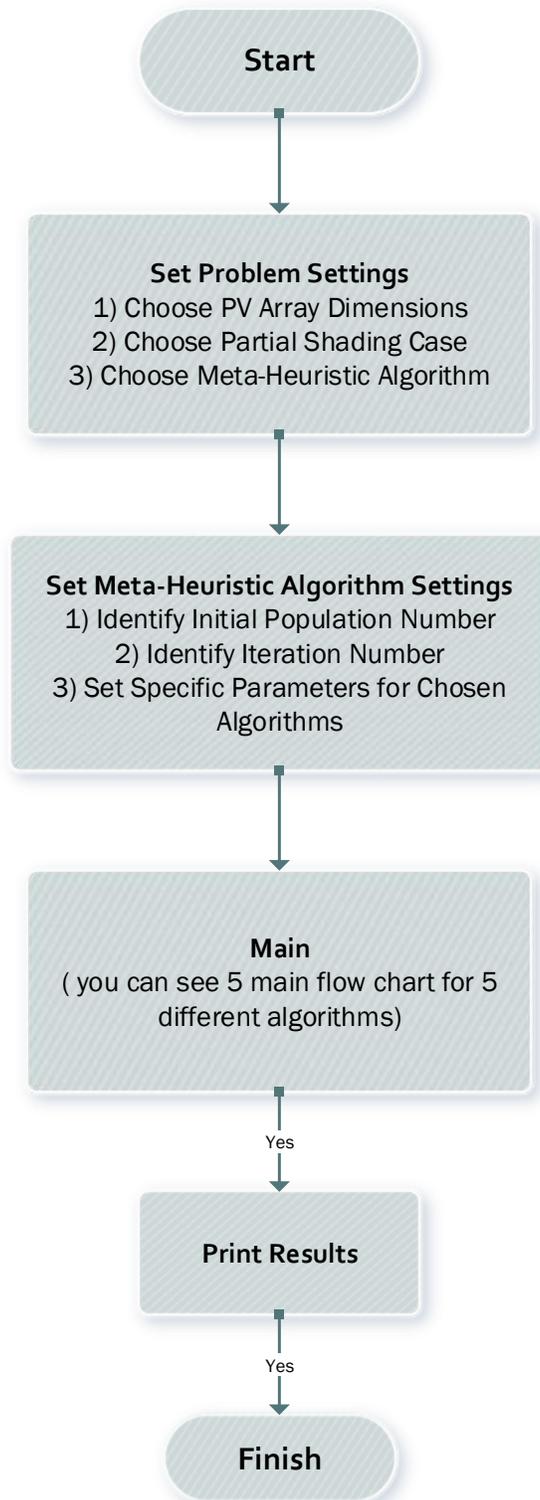

**Figure 6. general procedure of proposed algorithms**

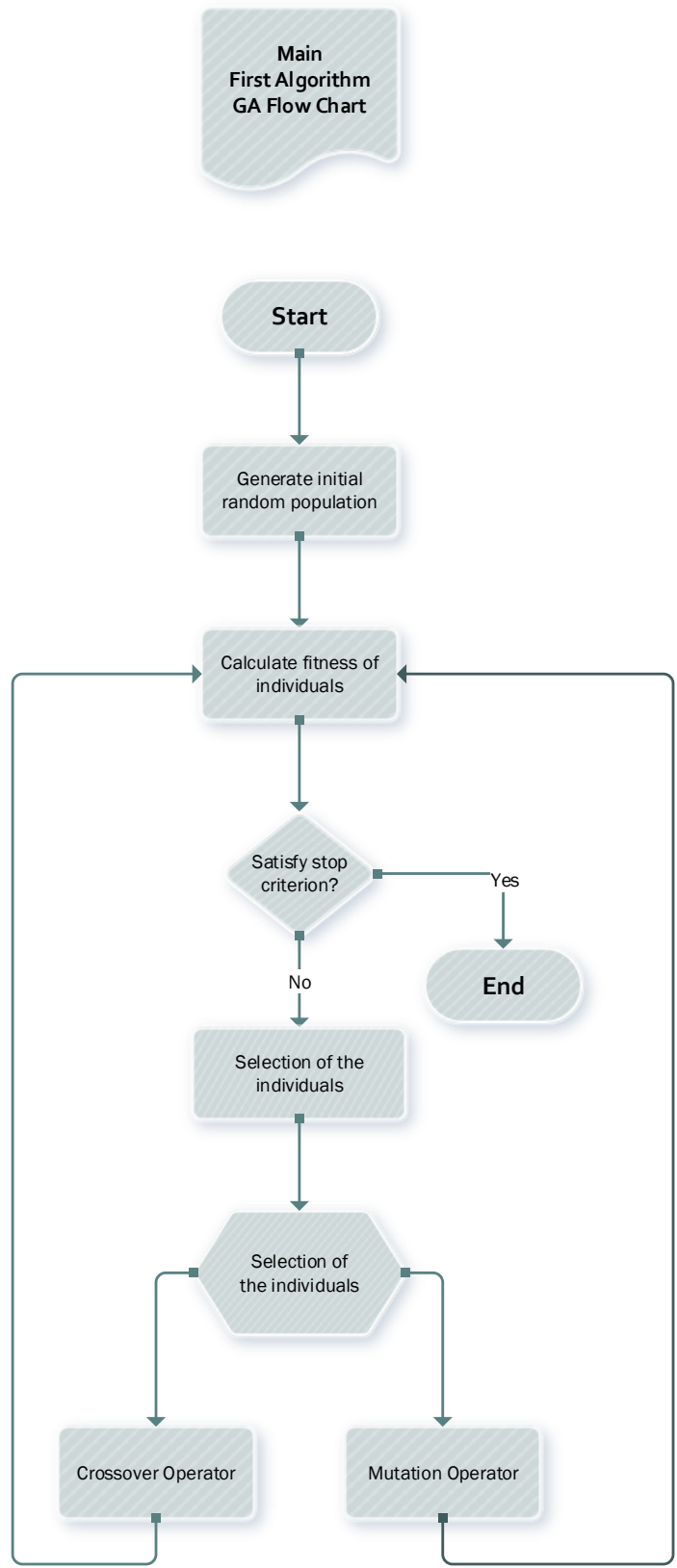

**Figure 7-a. Proposed Algorithms flowcharts (GA)**

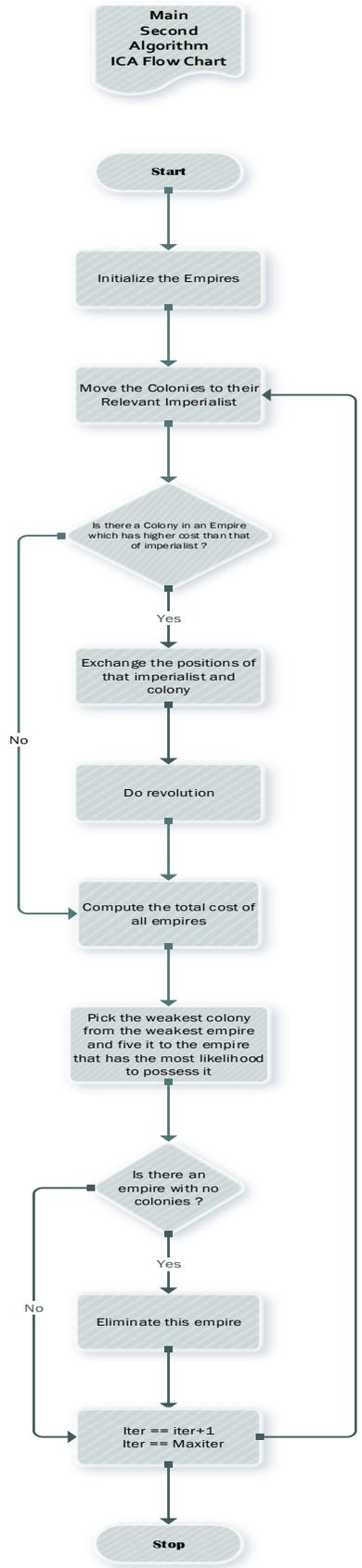

**Figure 7-*b*. Proposed Algorithms flowcharts (ICA)**

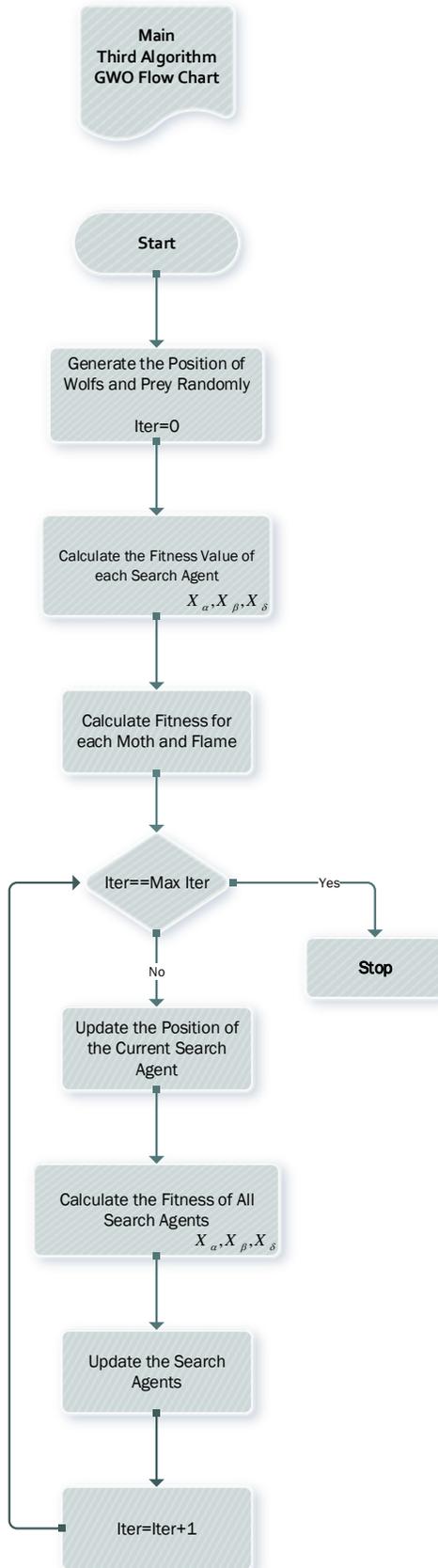

**Figure 7-*c*. Proposed Algorithms flowcharts (GWO)**

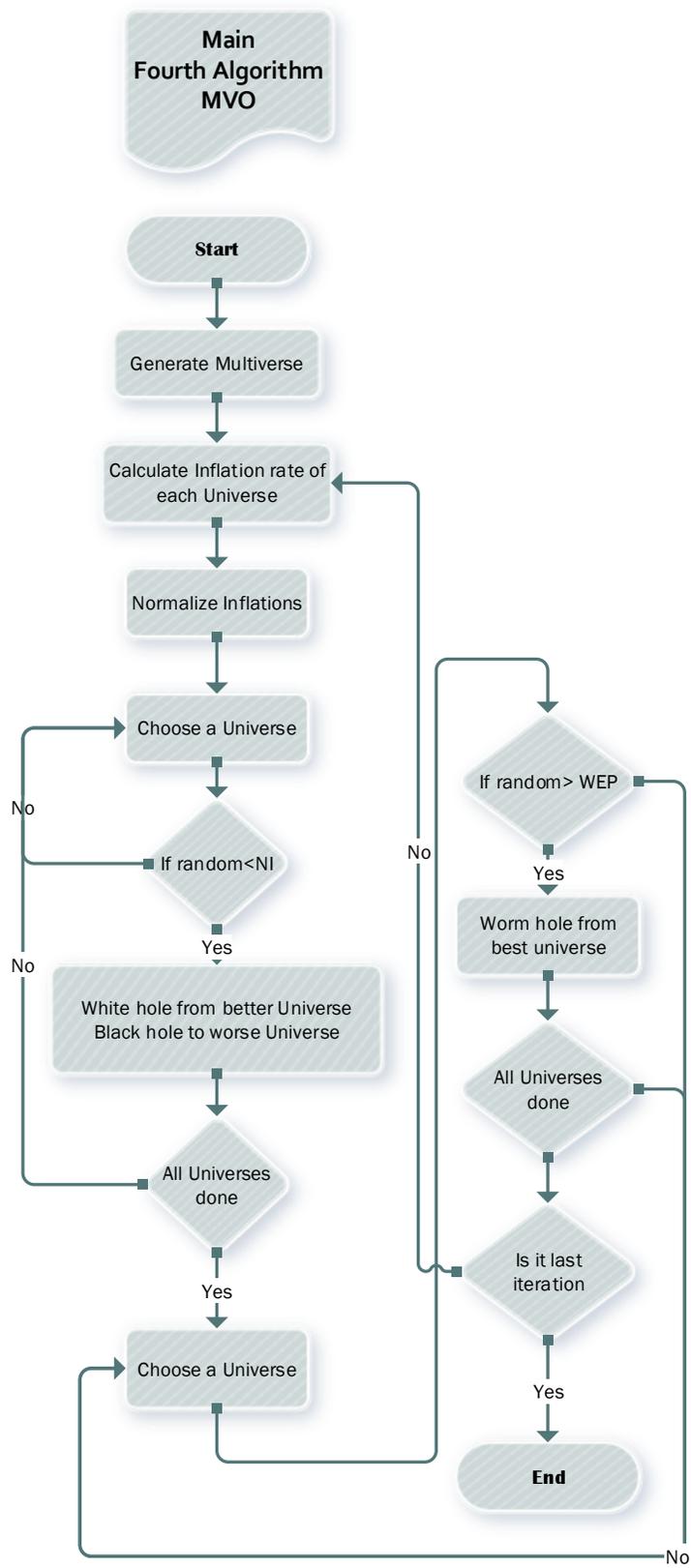

**Figure 7-d.** Proposed Algorithms flowcharts (MVO)

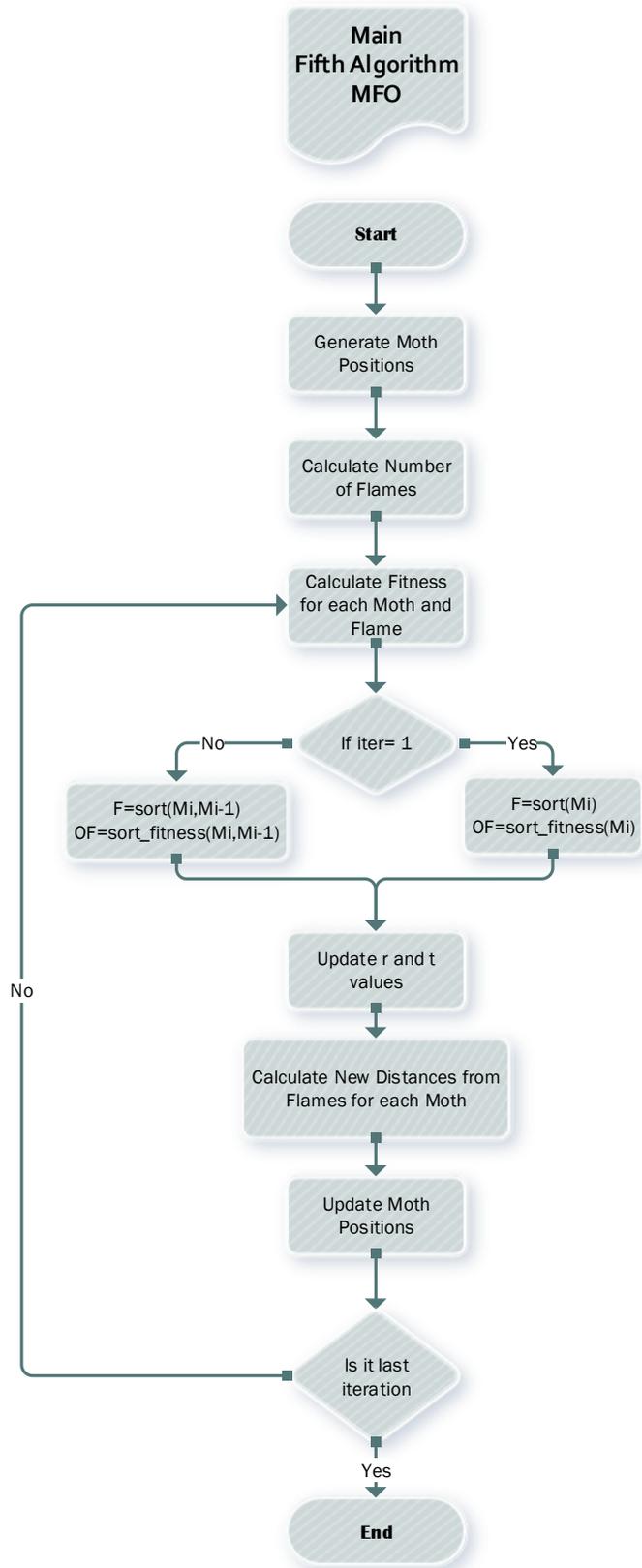

**Figure 7-e. Proposed Algorithms flowcharts (MFO)**

The basic procedure of proposed algorithms is shown in figure 6. First, a PV array with arbitrary dimensions is chosen, then one of four types of shading patterns and a metaheuristic algorithm from five algorithms described above are chosen. After that, we configure the initial parameters of metaheuristic algorithms. Choosing the right parameters for metaheuristic algorithms has a crucial effect on convergence to the best solution. In this paper, the parameters are chosen according to proposal parameters for each algorithm in their main papers and also considering the problem conditions. Moreover, some identical parameters are used to compare these algorithms. Therefore, the initial population number of each algorithm and the maximum number of iterations is set as 100 and 800, respectively for each run. Each algorithm is tested in 10 runs to be more confident with the results. As shown in figure 6, after choosing the parameters of the problem and metaheuristic algorithm, the chosen algorithm procedure will run as shown in figure 7's flowcharts for each algorithm.

## IV. EXPERIMENT RESULTS

As mentioned in session 2, four standard shading patterns are used to evaluate proposed metaheuristic algorithms for the defined problem. Also, a *9×9* TCT pattern is used to compare results with recent works. It is noteworthy to say that the proposed method can solve not only 9×9 but also *n×n* PV array problems. All five algorithms are considered on each shading pattern. The first experiment was about the minimum number of iterations required to find the best answer in each algorithm. In the next step, the average number of repetitions is being used to get the best response in each algorithm. Then, examining the required time to reach the maximum possible power for each algorithm. Finally, checking the correctness of the response of each algorithm. All algorithms are implemented and tested using simulation in MATLAB/Simulink environment.

### A. *First shading pattern (short-wide)*

The first shading pattern based on figure 8 is short-wide. As you can see almost half of the PV array is under shadings. The initial pattern consists of 4 different shadings with the irradiance of 900 *W/m²*, 600 *W/m²*, 400 *W/m²*, and 200 *W/m²*.

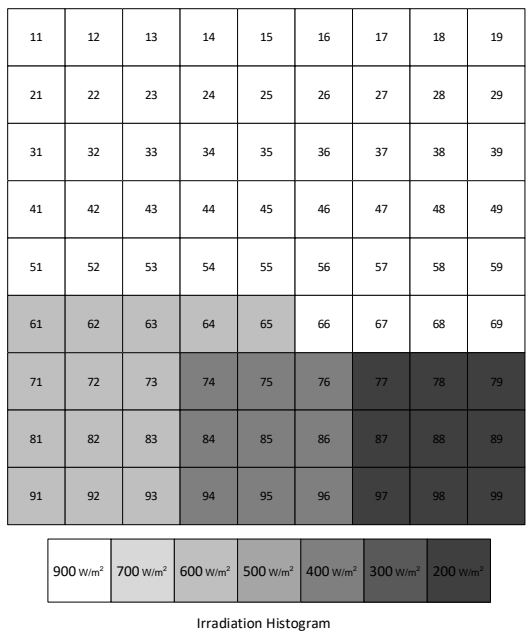

**Figure 8. TCT Configuration for shading pattern case 1**

For finding the location of GP, the row currents value must be calculated. The current generated by each row of PV array shown in figure 8 can be defined by using equation 11. For the shading pattern shown in Fig. 8, the cells in the rows 1 to 5 have the same insolation (900 $W/m^2$). Thus, the generated current of these rows is the same and is equal to $8.1I_m$ as shown in equation 16.

$$k_1 = k_2 = k_3 = k_4 = k_5 = 0.9 W/m^2$$
$$I_{R_1} = I_{R_2} = I_{R_3} = I_{R_4} = I_{R_5} = 9 \times k_1 \times I_m = 8.1 I_m \quad (17)$$

Due to 5 cells of the 6th row receive irradiance of 600 $W/m^2$, the generated current of this row can be defined from equation 11 as follows:

$$I_{R_6} = 5 \times 0.6 I_m + 4 \times 0.9 I_m = 6.6 I_m \quad (18)$$

The generated current of rows 7 to 9 is the same because they have the same pattern. Its value is defined as follows:

$$I_{R_7} = I_{R_8} = I_{R_9} = 3 \times 0.6 I_m + 3 \times 0.4 I_m + 3 \times 0.2 I_m = 3.6 I_m \quad (19)$$

As mentioned before, all modules of the same column connected in series to each other and all modules of each row connected in parallel to each other. Hence, the total power of the PV array is defined by equation 20.

$$P_{array} = I_{row} \times V_{array} \quad (20)$$

If the power requirement increases, the rows with the lowest current limits are bypassed. Hence, the total output power of the PV array is limited by the lowest current, and current of each row is limited by irradiance which is related to shading pattern on each row's cells. Table 3 shows the results of bypassing the limited current rows.

**Table 3. Location of GP in TCT Arrangement for Case 1**

| TCT Arrangement Results | | | |
|---|---|---|---|
| Row currents in order in which Panels are bypassed | | Voltage $V_{array}$ | Power $P_{array}$ |
| $I_{R9}$ | $3.6 I_m$ | $9 V_m$ | $32.4 V_m I_m$ |
| $I_{R8}$ | $3.6 I_m$ | - | - |
| $I_{R7}$ | $3.6 I_m$ | - | - |
| $I_{R6}$ | $6.6 I_m$ | $6 V_m$ | $39.6 V_m I_m$ |
| $I_{R5}$ | $8.1 I_m$ | $5 V_m$ | $40.5 V_m I_m$ |
| $I_{R4}$ | $8.1 I_m$ | - | - |
| $I_{R3}$ | $8.1 I_m$ | - | - |
| $I_{R2}$ | $8.1 I_m$ | - | - |
| $I_{R1}$ | $8.1 I_m$ | - | - |

Bypassing one or more rows of PV arrays is costly and difficult. Also, it may even not improve output power. The experiments show that maximum power obtained without bypassing any rows.

Now it is time to implement metaheuristic algorithms in Figure 7 to see what the result will be. By running all metaheuristic algorithms, the PV array shown in figure 9 was obtained. Experiments show that this arrangement is the best PV array arrangement for the first shading pattern, which gives us the maximum possible power.

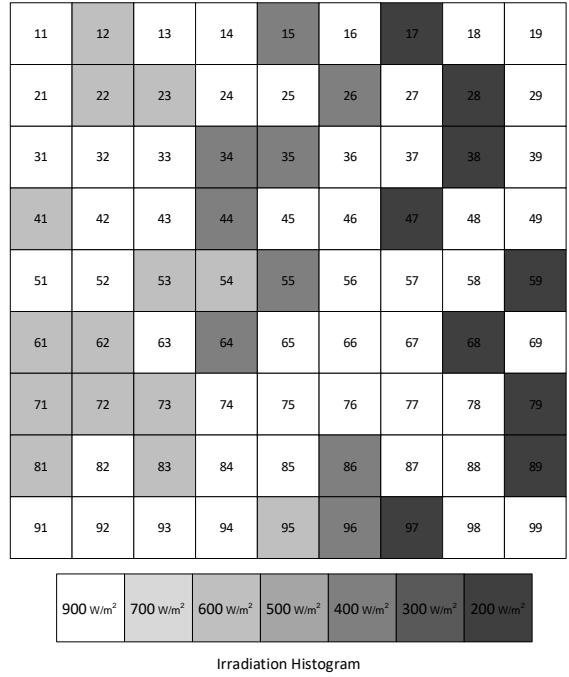

**Figure 9.** Meta-Heuristic Algorithms Configuration for shading pattern case 1

As previously described for computations of TCT configuration, the current of each row is calculated to evaluate the power of the PV array.

$$\begin{aligned}
I_{R_1} &= I_{R_4} = I_{R_9} = 6 \times 0.9 \times I_m + 0.6 \times I_m + 0.4 \times I_m + 0.2 \times I_m = 6.6 \, I_m \\
I_{R_2} &= I_{R_5} = I_{R_6} = I_{R_8} = 5 \times 0.9 \times I_m + 2 \times 0.6 \times I_m + 0.4 \times I_m + 0.2 \times I_m = 6.3 \, I_m \\
I_{R_3} &= 6 \times 0.9 \times I_m + 2 \times 0.4 \times I_m + 0.2 \times I_m = 6.6 \, I_m \\
I_{R_7} &= 5 \times 0.9 \times I_m + 3 \times 0.6 \times I_m + 0.2 \times I_m = 6.5 \, I_m
\end{aligned} \quad (20)$$

The power gains from the given arrangement defined in equation (21) is equal to $56.7 V_m I_m$.

$$P_{array} = 6.6 \, I_m \times 9 V_m = 56.7 V_m I_m \quad (21)$$

From the PV curve, it is concluded that this solution is 34.96% better than TCT arrangement. Furthermore, according to [10], it is 15.18% better than Sudoku algorithm. For showing the point that bypassing does not improve maximum power in this method, the power was also calculated by bypassing limited rows. Table 4 shows the results.

**Table 4. Location of GP in Meta-Heuristic Algorithms Arrangement for Case 1**

| Meta-heuristic Arrangement Results | | | |
|---|---|---|---|
| Row currents in order in which Panels are bypassed | | Voltage $V_{array}$ | Power $P_{array}$ |
| $I_{R9}$ | $6.3\ I_m$ | $9\ V_m$ | **$56.7\ V_m I_m$** |
| $I_{R8}$ | $6.3\ I_m$ | - | - |
| $I_{R7}$ | $6.3\ I_m$ | - | - |
| $I_{R6}$ | $6.3\ I_m$ | – | - |
| $I_{R5}$ | $6.4\ I_m$ | $5\ V_m$ | $32\ V_m I_m$ |
| $I_{R4}$ | $6.5\ I_m$ | $4\ V_m$ | $26\ V_m I_m$ |
| $I_{R3}$ | $6.6\ I_m$ | $3\ V_m$ | $19.8\ V_m I_m$ |
| $I_{R2}$ | $6.6\ I_m$ | - | - |
| $I_{R1}$ | $6.6\ I_m$ | - | - |

Maximum possible power is extracted without bypassing, unlike TCT arrangement as shown in table 4. For finding the best and fastest metaheuristic algorithms for case 1, first of all, algorithms are compared by the minimum number of iterations required to get the best answer. The results are shown in table 5.

**Table 5. Meta-Heuristic Algorithms Iteration Results for Case 1**

| Meta-heuristic Algorithms Iteration Results | | | | |
|---|---|---|---|---|
| Algorithms Names | Minimum Iteration Require to Reach Best Power | Mean Iteration Require to Reach Best Power | Rank based Best Iteration | Rank based Mean Iteration |
| GA | **9** | **29** | **1** | **1** |
| ICA | 59 | 140 | 3 | 3 |
| GWO | 111 | 333 | 4 | 5 |
| MFO | 27 | 43 | 2 | 2 |
| MVO | 152 | 263 | 5 | 4 |

Table 5 columns are the minimum number of iterations required to reach best output power, the mean number of iterations required to get the best output power and algorithm's rankings based on the minimum and mean required iterations to get the best output power, respectively. As shown, the best algorithm for case 1 is the Genetic algorithm.

As mentioned before, each algorithm is run for 800 iterations, but these algorithms need less than 800 iterations to reach the best answer. Therefore, the minimum required time is less than table 6 values.

Table 6. Meta-Heuristic Algorithms Time Results for Case 1

**Meta-heuristic Algorithms Time Results**

| Algorithms Names | Mean time For 800 Iterations | Rank based Mean Iteration |
|---|---|---|
| GA | 11.9971 | 4 |
| ICA | 10.3436 | 3 |
| GWO | **4.5448** | 1 |
| MFO | 22.0201 | 5 |
| MVO | 10.9953 | 2 |

Table 6 shows the mean required time for 800 iterations for the first case. It is clear that GWO algorithm has the fastest performance (4.5 seconds) and MFO algorithm is the slowest one. Algorithms are ranked based on the minimum required time in the third column. According to table 5 and 6 for evaluating the best algorithm, equation 22 is used.

$$Mean\ Time = \frac{Mean\ time\ for\ 800\ iteration}{800} \times Mean\ iteration\ require\ to\ reach\ the\ best\ power \quad (22)$$

The results show in table 7.

Table 7. Best Meta-Heuristic Algorithms for Case 1

**Best Meta-heuristic Algorithms for Case 1**

| Algorithms Names | Mean time For 800 Iterations | Rank based Mean Iteration |
|---|---|---|
| **GA** | **0.4348** | 1 |
| ICA | 1.8101 | 3 |
| GWO | 1.8917 | 4 |
| MFO | 1.1835 | 2 |
| MVO | 3.6147 | 5 |

The results show that the Genetic algorithm has the best performance and MVO is the worst. Another noteworthy point is that the mean time of all algorithms is less than 2 seconds, which indicates that metaheuristic algorithms are good solutions to the problem.

### B. *Second shading pattern (long-narrow)*

The second shading pattern shown in figure 10 is long-narrow. As you can see almost one-fifth of PV cells in the PV array receive irradiance of 600 $W/m^2$, 400 $W/m^2$, and 300 $W/m^2$.

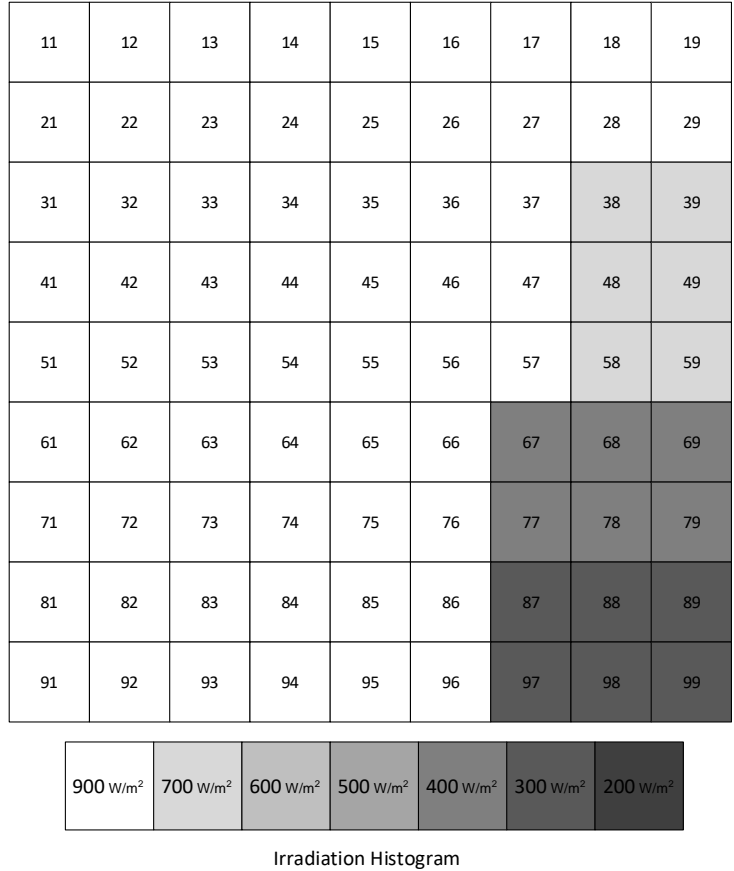

**Figure 10. TCT Configuration for shading pattern case 2**

Each current of rows and total PV array power calculate according to equations described in first shading patterns. We expect more power than previous patterns, because there are fewer cells affected by shadings. The results in table 8 confirm our expectation.

**Table 8. Location of GP in TCT Arrangement for Case 2**

| TCT Arrangement Results | | | |
|---|---|---|---|
| Row currents in order in which Panels are bypassed | | Voltage $V_{array}$ | Power $P_{array}$ |
| $I_{R9}$ | $3.6\ I_m$ | $9\ V_m$ | $56.7\ V_m I_m$ |
| $I_{R8}$ | $3.6\ I_m$ | - | - |
| $I_{R7}$ | $3.6\ I_m$ | $7\ V_m$ | $46.2\ V_m I_m$ |
| $I_{R6}$ | $6.6\ I_m$ | – | - |
| $I_{R5}$ | $8.1\ I_m$ | $5\ V_m$ | $38.5\ V_m I_m$ |
| $I_{R4}$ | $8.1\ I_m$ | - | - |
| $I_{R3}$ | $8.1\ I_m$ | - | - |
| $I_{R2}$ | $8.1\ I_m$ | $2\ V_m$ | $16.2\ V_m I_m$ |
| $I_{R1}$ | $8.1\ I_m$ | - | - |

Table 9 shows the result of implementing metaheuristic algorithms on the PV array shown in figure 10. As mentioned earlier, metaheuristic algorithms assure us to extract maximum power without bypassing PV arrays. Figure 11 shows the obtained PV array using metaheuristic algorithms.

**Table 9. Location of GP in Meta-Heuristic Algorithms Arrangement for Case 2**

| Meta-heuristic Arrangement Results | | | |
|---|---|---|---|
| Row currents in order in which Panels are bypassed | | Voltage $V_{array}$ | Power $P_{array}$ |
| $I_{R9}$ | $7.1\ I_m$ | $9\ V_m$ | $63.9 V_m I_m$ |
| $I_{R8}$ | $7.1\ I_m$ | - | - |
| $I_{R7}$ | $7.1\ I_m$ | - | - |
| $I_{R6}$ | $7.1\ I_m$ | – | - |
| $I_{R5}$ | $7.3\ I_m$ | $5\ V_m$ | $36.5\ V_m I_m$ |
| $I_{R4}$ | $7.3\ I_m$ | - | - |
| $I_{R3}$ | $7.3\ I_m$ | - | - |
| $I_{R2}$ | $7.3\ I_m$ | - | - |
| $I_{R1}$ | $7.5\ I_m$ | $1\ V_m$ | $7.5 V_m I_m$ |

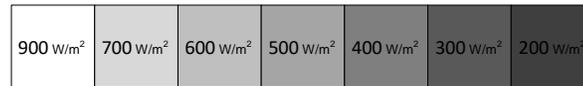

Irradiation Histogram

**Figure 11. Meta-Heuristic Algorithms Configuration for shading pattern case 2**

Comparing the results with recent methods show that the proposed method has 7.8% enhancement towards Sudoku configuration and 1.93% towards TCT configuration. Comparing the five proposed algorithms shown in table 10 to find the best metaheuristic algorithm for second shading pattern.

Table 10 shows the minimum number of iterations required to generate the best output power for each algorithm in the second case. It also has a mean of required iterations in the third column. Last two columns contain ranking of algorithms. It is clear that in spite of the first case, GWO has the best score and similar to the first case MVO is the worst.

Table 10. Meta-Heuristic Algorithms Iteration Results for Case 2

**Meta-heuristic Algorithms Iteration Results**

| Algorithms Names | Minimum Iteration Require to Reach Best Power | Mean Iteration Require Reach Best Power | Rank based Best Iteration | Rank based Mean Iteration |
|---|---|---|---|---|
| GA | 78 | 144 | 4 | 2 |
| ICA | 74 | 315 | 3 | 5 |
| GWO | **18** | 200 | **1** | 3 |
| MFO | 32 | **47** | 2 | **1** |
| MVO | 201 | 217 | 5 | 4 |

Required time for 800 iterations shown in table 11. The results show that GWO is the fastest and MFO is the slowest one.

Table 11. Meta-Heuristic Algorithms Time Results for Case 2

**Meta-heuristic Algorithms Time Results**

| Algorithms Names | Mean time For 800 Iterations | Rank based Mean Iteration |
|---|---|---|
| GA | 11.6729 | 4 |
| ICA | 8.3405 | 2 |
| GWO | **4.506** | **1** |
| MFO | 21.1248 | 5 |
| MVO | 10.1031 | 3 |

Also, the best algorithm for this arrangement is chosen according to the details mentioned in the first case.

Table 12. Best Meta-Heuristic Algorithms for Case 2

**Best Meta-heuristic Algorithms for Case 2**

| Algorithms Names | Mean time For 800 Iterations | Rank based Mean Iteration |
|---|---|---|
| GA | 2.1011 | 3 |
| ICA | 3.2840 | 5 |
| **GWO** | **1.1265** | **1** |
| MFO | 1.2411 | 2 |
| MVO | 2.7404 | 4 |

As a result, GWO and MFO are the best algorithms to solve the second case of partial shading pattern according to table 12. Another important point is that the mean required time to reach the best power of all algorithms is 0.2 seconds more than the first case, indicating the difficulty of this case.

## C. Third shading pattern (short-narrow)

This pattern shown in figure 12 is very rare in PV arrays. In this pattern, fewer cells are affected by shadings. In the standard test array, there are two irradiances of 400 $W/m^2$ and 600 $W/m^2$.

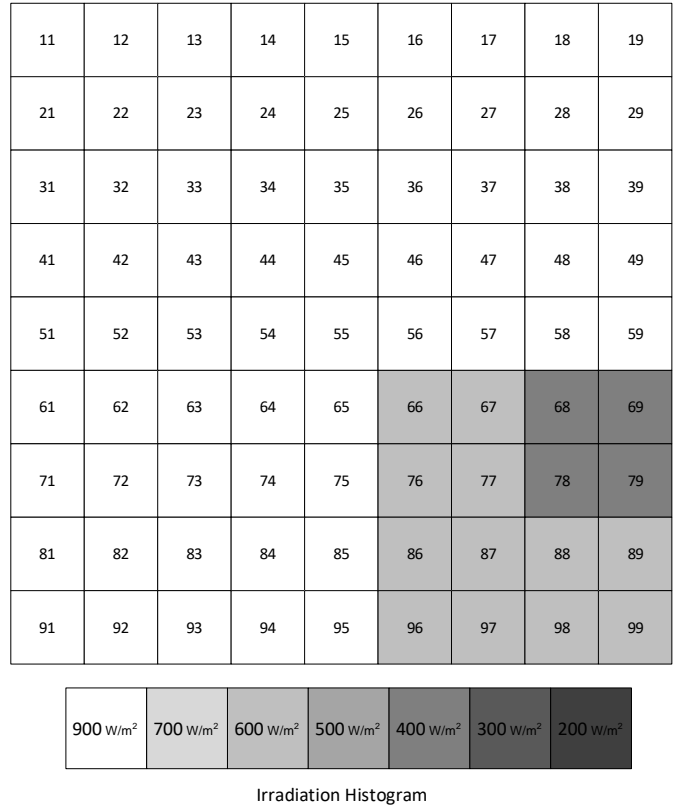

Figure 12. TCT Configuration for shading pattern case 3

The currents and power of the PV array calculate according to previous patterns. Table 13 shows the results. Table 14 shows the results of metaheuristic algorithms implemented on the PV array shown in figure 12. Similar to previous shading patterns there is no need to bypass limited current rows to extract maximum possible output power in this case too.

Table 13. Location of GP in TCT Arrangement for Case 3

| TCT Arrangement Results | | | |
|---|---|---|---|
| Row currents in order in which Panels are bypassed | | Voltage $V_{array}$ | Power $P_{array}$ |
| $I_{R9}$ | 6.1 $I_m$ | 9 $V_m$ | 54.9 $V_m I_m$ |
| $I_{R8}$ | 6.1 $I_m$ | - | - |
| $I_{R7}$ | 7.3 $I_m$ | 7 $V_m$ | 51.1 $V_m I_m$ |
| $I_{R6}$ | 7.3 $I_m$ | – | - |
| $I_{R5}$ | 8.1 $I_m$ | 5 $V_m$ | 40.5 $V_m I_m$ |
| $I_{R4}$ | 8.1 $I_m$ | - | - |
| $I_{R3}$ | 8.1 $I_m$ | - | - |
| $I_{R2}$ | 8.1 $I_m$ | - | - |
| $I_{R1}$ | 8.1 $I_m$ | - | - |

**Table 14. Location of GP in Meta-Heuristic Algorithms Arrangement for Case 3**

## Meta-heuristic Arrangement Results

| Row currents in order in which Panels are bypassed | | Voltage $V_{array}$ | Power $P_{array}$ |
|---|---|---|---|
| $I_{R9}$ | 7.3 $I_m$ | 9 $V_m$ | **65.7$V_m I_m$** |
| $I_{R8}$ | 7.3 $I_m$ | - | - |
| $I_{R7}$ | 7.5 $I_m$ | 7 $V_m$ | 52.5 $V_m I_m$ |
| $I_{R6}$ | 7.5 $I_m$ | – | - |
| $I_{R5}$ | 7.5 $I_m$ | - | - |
| $I_{R4}$ | 7.5 $I_m$ | - | - |
| $I_{R3}$ | 7.5 $I_m$ | - | - |
| $I_{R2}$ | 7.6 $I_m$ | 2 $V_m$ | 15.2 $V_m I_m$ |
| $I_{R1}$ | 7.6 $I_m$ | - | - |

Figure 13 shows the PV array resulted by metaheuristic algorithms that have maximum possible power.

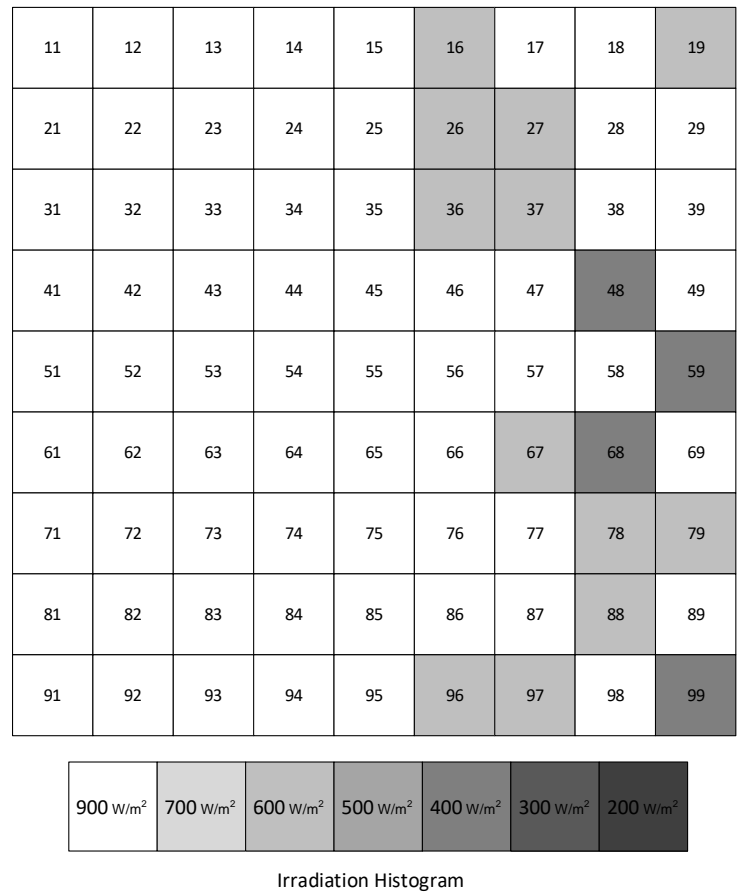

Irradiation Histogram

**Figure 13. Meta-Heuristic Algorithms Configuration for shading pattern case 3**

The proposed result is 7.28 percent better than TCT arrangement and 1.42 percent better than Sudoku arrangement. The number of required iterations for maximum output power shown in table 15. GWO, ICA, and GA have the best answers.

**Table 15. Meta-Heuristic Algorithms Iteration Results for Case 3**

| Algorithms Names | Minimum Iteration Require to Reach Best Power | Mean Iteration Require Reach Best Power | Rank based Best Iteration | Rank based Mean Iteration |
|---|---|---|---|---|
| GA | 1 | 1 | 1 | 1 |
| ICA | 1 | 1 | 1 | 1 |
| GWO | 1 | 1 | 1 | 1 |
| MFO | 1 | 2 | 1 | 2 |
| MVO | 1 | 5 | 1 | 3 |

According to table 15, if the problem is simple, metaheuristic algorithms can find the best solution in the first iteration. Therefore, the metaheuristic algorithms are very fast and appropriate for PV array problem. The required time for 800 iterations is shown in table 16.

**Table 16. Meta-Heuristic Algorithms Time Results for Case 3**

| Algorithms Names | Mean time for 800 Iterations | Rank based Mean Iteration |
|---|---|---|
| GA | 11.2674 | 4 |
| ICA | 8.9325 | 2 |
| GWO | **4.5546** | **1** |
| MFO | 21.1248 | 5 |
| MVO | 10.1031 | 3 |

Required time to reach the maximum possible power for each algorithm is calculated by equation 22.

**Table 17. Best Meta-Heuristic Algorithms for Case 3**

| Algorithms Names | Mean time For 800 Iterations | Rank based Mean Iteration |
|---|---|---|
| GA | 0.0141 | 3 |
| ICA | 0.0111 | 2 |
| **GWO** | **0.0056** | **1** |
| MFO | 0.0528 | 4 |
| MVO | 0.0631 | 5 |

Mean required time for this pattern is less than 0.03 seconds. This shows that metaheuristic algorithms have a great performance on the problem. GWO has the best result and MVO has the worst result for this pattern.

## D. Forth shading pattern (long-wide)

The last pattern is long-wide. About 63 percent of cells in a PV array receive irradiance of 600 $W/m^2$, 500 $W/m^2$, 400 $W/m^2$, and 200 $W/m^2$. Therefore, it is expected to get less power compared to other patterns. Figure 14 shows the PV array pattern.

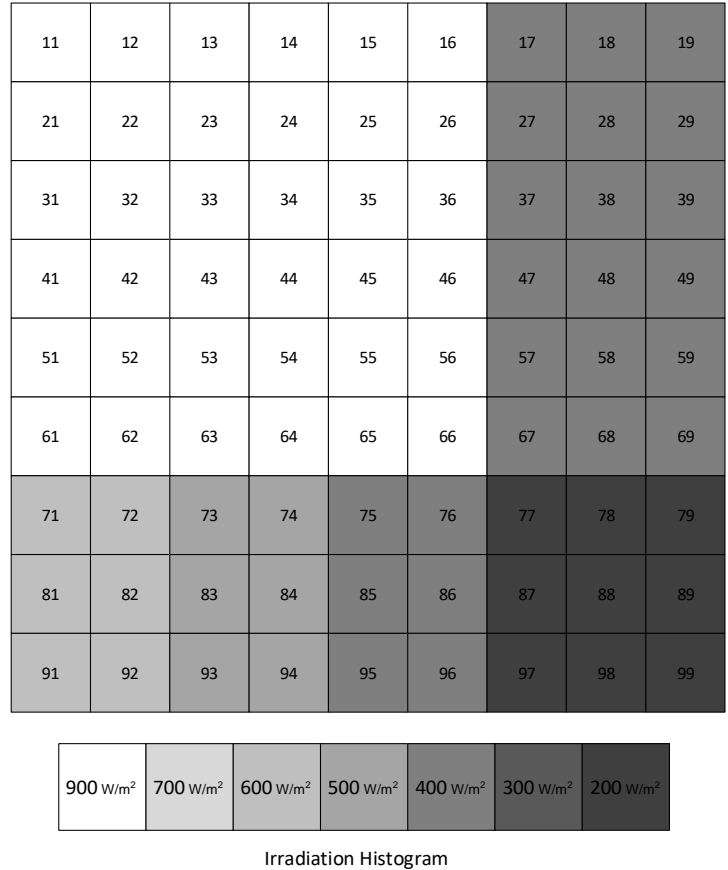

Figure 14. TCT Configuration for shading pattern case 4

Table 18 shows PV array output power.

Table 18. Location of GP in TCT Arrangement for Case 4

| TCT Arrangement Results | | | |
|---|---|---|---|
| Row currents in order in which Panels are bypassed | | Voltage $V_{array}$ | Power $P_{array}$ |
| $I_{R9}$ | 6.1 $I_m$ | 9 $V_m$ | 32.4 $V_m I_m$ |
| $I_{R8}$ | 6.1 $I_m$ | - | - |
| $I_{R7}$ | 7.3 $I_m$ | - | - |
| $I_{R6}$ | 7.3 $I_m$ | 6 $V_m$ | 39.6 $V_m I_m$ |
| $I_{R5}$ | 8.1 $I_m$ | - | - |
| $I_{R4}$ | 8.1 $I_m$ | - | - |
| $I_{R3}$ | 8.1 $I_m$ | - | - |
| $I_{R2}$ | 8.1 $I_m$ | - | - |
| $I_{R1}$ | 8.1 $I_m$ | - | - |

Figure 15 shows the PV array arrangement after using metaheuristic algorithms.

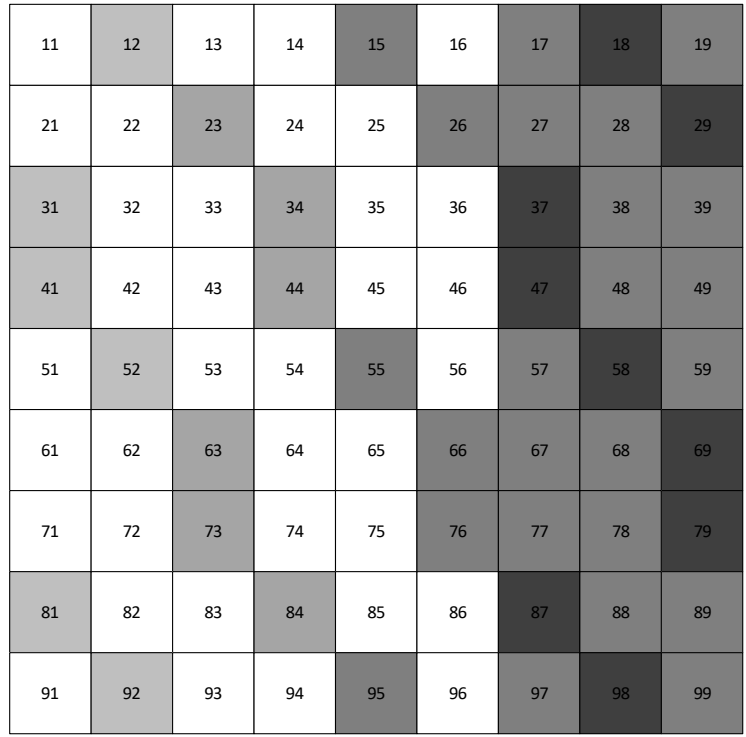

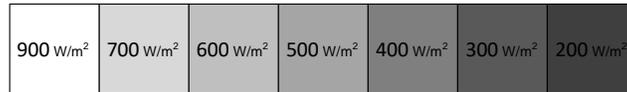

Irradiation Histogram

**Figure 15. Meta-Heuristic Algorithms Configuration for shading pattern case 4**

Table 19 shows that reaching maximum possible power can be possible without bypassing the current rows.

**Table 19. Location of GP in Meta-Heuristic Algorithms Arrangement for Case 4**

| Meta-heuristic Arrangement Results | | | |
|---|---|---|---|
| Row currents in order in which Panels are bypassed | | Voltage $V_{array}$ | Power $P_{array}$ |
| $I_{R9}$ | 5.5 $I_m$ | 9 $V_m$ | **49.5$V_m I_m$** |
| $I_{R8}$ | 5.5 $I_m$ | - | - |
| $I_{R7}$ | 5.5 $I_m$ | - | - |
| $I_{R6}$ | 5.6 $I_m$ | 6 $V_m$ | 33.6 $V_m I_m$ |
| $I_{R5}$ | 5.6 $I_m$ | - | - |
| $I_{R4}$ | 5.6 $I_m$ | - | - |
| $I_{R3}$ | 5.7 $I_m$ | 3 $V_m$ | 17.1 $V_m I_m$ |
| $I_{R2}$ | 5.7 $I_m$ | - | - |
| $I_{R1}$ | 5.7 $I_m$ | - | - |

Enhancement of 24.09 percent towards TCT arrangement and 1.2 percent towards Sudoku arrangement prove that the proposed method performs better than other methods. According to table 20, the Genetic algorithm has the best performance on case number four, but MFO is the fastest one.

**Table 20. Meta-Heuristic Algorithms Iteration Results for Case 4**

| Algorithms Names | Minimum Iteration Require to Reach Best Power | Mean Iteration Require Reach Best Power | Rank based Best Iteration | Rank based Mean Iteration |
|---|---|---|---|---|
| GA | **5** | 25 | **1** | 2 |
| ICA | 8 | 101 | 3 | 3 |
| GWO | 9 | 272 | 4 | 5 |
| MFO | 6 | **29** | 2 | **1** |
| MVO | 141 | 220 | 5 | 4 |

Mean required time for 800 iterations of all algorithms are shown in table 21.

**Table 21. Meta-Heuristic Algorithms Time Results for Case 4**

| Algorithms Names | Mean time for 800 Iterations | Rank based Mean Iteration |
|---|---|---|
| GA | 12.2571 | 4 |
| ICA | 9.7342 | 2 |
| GWO | **4.7125** | **1** |
| MFO | 22.0999 | 5 |
| MVO | 11.0914 | 3 |

At last for finding the best algorithm for forth arrangement, the required time of each algorithm was calculated according to equation 22.

**Table 22. Best Meta-Heuristic Algorithms for Case 4**

| Algorithms Names | Mean time for 800 Iterations | Rank based Mean Iteration |
|---|---|---|
| GA | **0.3830** | **1** |
| ICA | 1.2289 | 3 |
| GWO | 1.6022 | 4 |
| MFO | 0.8011 | 2 |
| MVO | 3.0501 | 5 |

According to table 22, mean required time to reach the maximum possible power for this arrangement is 1.4136 seconds. As compared to the first arrangement (short-wide) the time is almost the same. The best and worst algorithms, in this case, are Genetic and MVO respectively.

*E. General Analysis*

Experiments were done using proposed metaheuristic algorithms on four different partial shading PV arrays. The results show that the proposed algorithms are superior to previous methods. Also, some analyses were done to find the fastest metaheuristic algorithm in all cases. The results are shown in figure 16.

**Figure 16. Mean required time to reach the maximum possible power of different shading patterns**

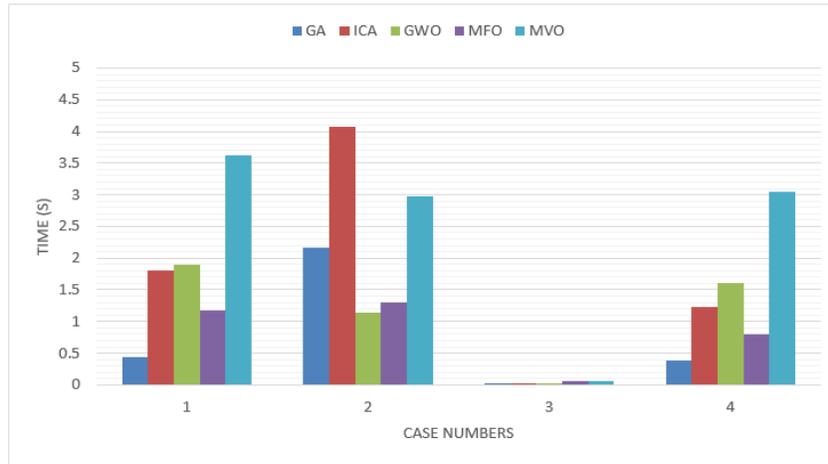

By looking precisely in figure 16, it is concluded that the most complex shading patterns like the second case (long-narrow), the worst needed time for metaheuristic algorithms to achieve the best answer is about 4 seconds. This shows that even in the hardest situations of PV arrays metaheuristic algorithms are an excellent choice to solve the problem. Also, if shading patterns are simpler like in the third case, metaheuristic algorithms achieve the best solution in less than 0.05 seconds.

As a result, in long shading patterns like long-narrow and long-wide, GWO is the fastest algorithm, and in short shading patterns like short-narrow and short-wide, the Genetic algorithm finds the solution in the lowest possible time. ICA was the worst in the second shading pattern, and MVO was the worst in the other three shading patterns. Figure 17 shows more detail information about each algorithm performance.

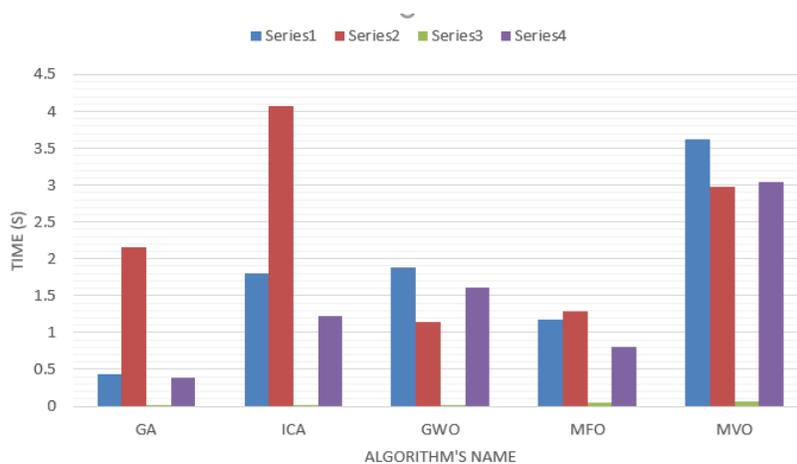

**Figure 17. Mean required time of proposed algorithms**

According to figure 17, mean required time of all shading patterns for each algorithm is less than 4.1 seconds. Genetic algorithm, MFO, and GWO are in first, second and third ranks of fastest algorithms. MVO has the worst performance among all algorithms.

The convergence curves shown in figure 18 to 21 show the best speed of all algorithms in each case. In the first case, GA is the fastest, but MVO has reached the maximum possible power in iteration 153 and later than other algorithms.

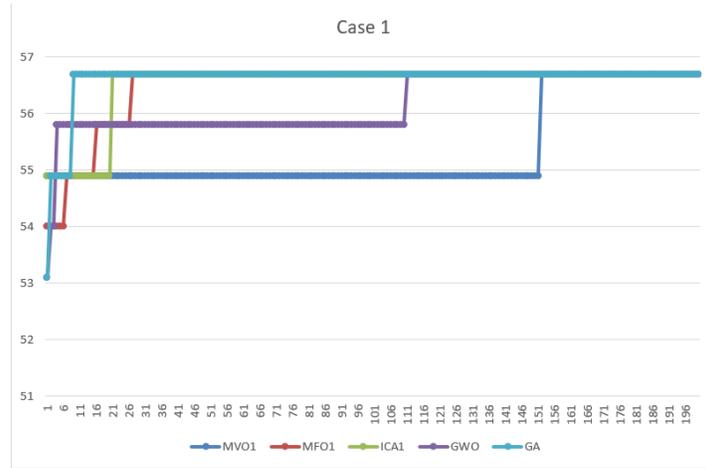

**Figure 18. Convergence curves case 1**

In the second case which is more difficult, GWO has obtained the best result faster than others.

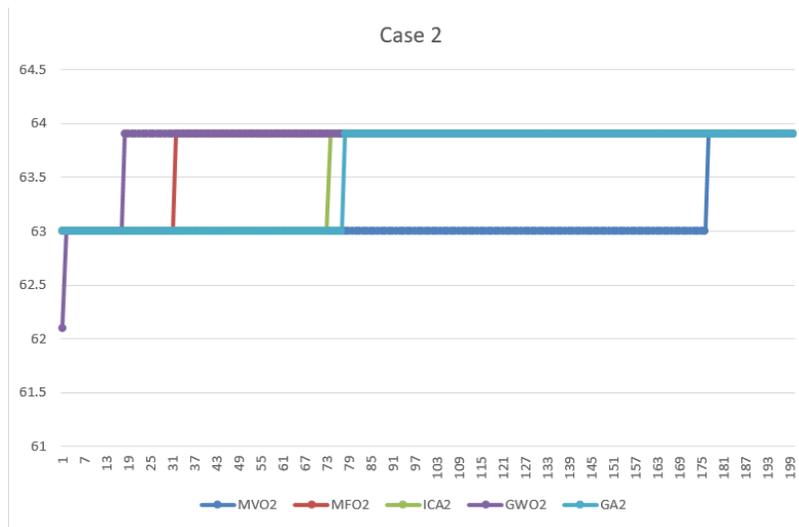

**Figure 19. Convergence curves case 2**

In the third case which is much simpler than the others, all algorithms obtained the best answer in the first iteration.

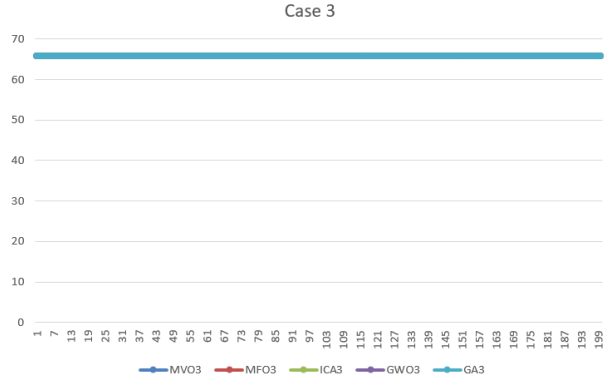

**Figure 20. Convergence curves case 3**

At last, in the fourth case Genetic algorithm is the fastest similar to the first case.

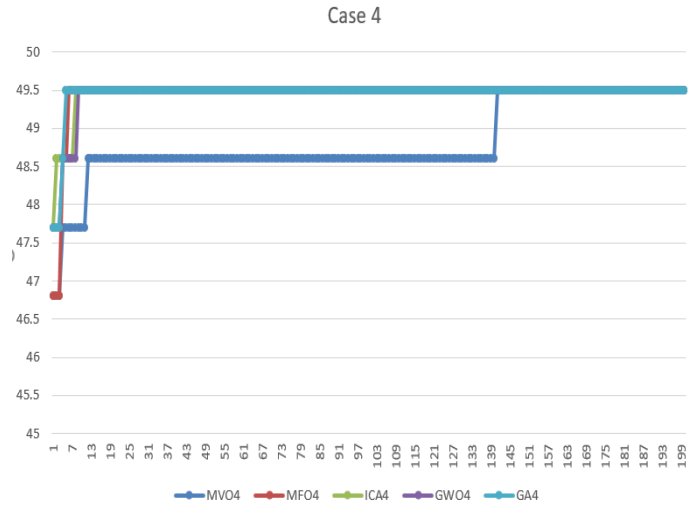

**Figure 21. Convergence curves case 4**

One of the main challenges in using metaheuristic algorithms is being stuck in local minimums, causing the algorithm are not able to find the global optimum. Hence, evaluating the metaheuristic algorithms based on their correctness of the result is required. In order to measure how many times each algorithm finds the global optimum answer; each algorithm runs 10 times in every shading pattern. The results are shown in table 23.

**Table 23. Meta-Heuristic Algorithms Correctness Results**

| Meta-heuristic Algorithms Correctness Results | | | | | | |
|---|---|---|---|---|---|---|
| Algorithms Names | Case 1 | Case 2 | Case 3 | Case 4 | Mean | Rank |
| GA | 100 | 30 | 100 | 80 | 77.5 | 4 |
| ICA | 100 | 90 | 100 | 80 | 92.5 | 2 |
| GWO | **100** | **100** | **100** | **100** | **100** | **1** |
| MFO | 70 | 70 | 100 | 90 | 82.5 | 3 |
| MVO | 50 | 40 | 100 | 40 | 57.5 | 5 |
| Case Means | 84 | 66 | 100 | 78 | 82 | - |

The first column of table 23 shows the name of each metaheuristic algorithm. The percentage of correctness in 10 times running of each algorithm is shown in columns 2 to 5 for each case. The sixth column shows the mean of the correctness of each algorithm and at last in the final column algorithms are ranked based on their correctness in finding the best solution. GWO is the only algorithm which has obtained maximum possible power in all cases. An important note is that mean of all algorithms correctness in all different cases is 82%. Also, the MVO does not have good results here as in previous experiments. Thus, MVO is not an appropriate algorithm for partial shading pattern problems. Although the genetic algorithm proposed in paper [10] was improved up to 17.5%, it ranked fourth among other algorithms. The mean percentage of correctness of metaheuristic algorithms under different shading patterns shown in the last row of table 23. The best solution is obtained in the third case and the worst one in the second case. Consequently, according to results in table 23 and figures 16 and 17 GWO can be chosen as the best algorithm for partial shading PV arrays problem due to its 100% correctness in all cases, best convergence time in second and third cases and reasonable convergence time in first and fourth cases.

## V. CONCLUSION

In this paper, the authors tried to find a method to extract maximum possible power in the shortest time in the photovoltaic cells under partial shading condition. For this purpose, five different algorithms were chosen from four groups of metaheuristic algorithms. These algorithms achieve the highest possible output power by changing the electrical connections between PV modules while their physical location is fixed. Thus, the proposed method achieves distributed shades on PV array and prevents concentrating the shadings on one location. Also, in order to get more power, there is no need to bypass any limited current rows. The required time for each algorithm to reach the best answer and correctness percentage is analysed. The results show that considering shading patterns on PV arrays; different algorithms can obtain the best answer. However, by examining the accuracy of all proposed algorithms to find the best solution, GWO was chosen as the best algorithm for solving the problem.